\documentclass[11pt]{article}
\usepackage[T1]{fontenc}
\usepackage[utf8]{inputenc}
\usepackage{lmodern,microtype}
\usepackage[a4paper,margin=1in]{geometry}
\usepackage{setspace}
\usepackage{amsmath,amssymb}
\usepackage{graphicx}
\usepackage{booktabs,array,tabularx,multirow}
\usepackage[labelfont=bf,font=small,labelsep=period]{caption}
\usepackage[super,numbers,sort&compress]{natbib}
\usepackage[most]{tcolorbox}
\usepackage{xcolor}
\usepackage{titlesec,needspace}
\usepackage{enumitem}
\usepackage[hidelinks]{hyperref}
\definecolor{navy}{HTML}{355E7C}
\definecolor{coral}{HTML}{BF7B40}
\definecolor{teal}{HTML}{2A9D8F}
\definecolor{paleblue}{HTML}{E8F0F5}
\definecolor{ink2}{HTML}{5B6069}

\titleformat{\section}{\sffamily\bfseries\large}{}{0pt}{}
\titleformat{\subsection}{\sffamily\bfseries\normalsize}{}{0pt}{}
\titlespacing*{\section}{0pt}{14pt}{6pt}
\titlespacing*{\subsection}{0pt}{10pt}{3pt}
\setlist{nosep,leftmargin=*}

\newcounter{pbox}
\newtcolorbox{perspectivebox}[2][]{enhanced,float=tp,colback=paleblue,colframe=navy,boxrule=0.6pt,arc=2pt,
  left=8pt,right=8pt,top=6pt,bottom=6pt,fonttitle=\sffamily\bfseries,coltitle=white,colbacktitle=navy,
  before upper={\refstepcounter{pbox}\setlength{\parskip}{4pt}},title={#2},#1}

\newcommand{\rebecaref}{\cite{yan2026rebeca}}
\newcommand{\co}[1]{\mathrm{CO}_{#1}}
\newcommand{\spc}[1]{\mathrm{Sp}_{#1}}

\title{\sffamily\bfseries\LARGE Behaviour analysis for large language model-driven agents}
\author{%
  \sffamily Tianqiang Yan$^{\dag}$ and Lizhen Qu$^{\dag,\ast}$\\[4pt]
  \sffamily\small $^{\dag}$Faculty of Information Technology, Monash University, Melbourne, Australia\\
  \sffamily\small $^{\ast}$Corresponding author: lizhen.qu@monash.edu%
}
\date{}

\begin{document}
\maketitle
\begin{abstract}
\noindent\sffamily
Understanding language-agent behaviour requires learning how actions are organised across a trajectory and how this organisation relates to context and outcomes. We propose a general framework that connects data-driven behavioural discovery with analysis and experimental investigation. Categories developed from recorded activity define what is measured. Analyses preserve the behaviours' positions in the process and express their relationships as interpretable hypotheses. Statistical assessment and tests in held-out contexts examine the strength and scope of the evidence. Experiments, including prompt steering, then test how behaviour and outcomes change under deliberately varied conditions. A study of iterative self-reflection illustrates how discovery, dependency learning and prompt comparisons can inform one another. We develop this approach for trajectories of varying length, event order and agent roles, and identify advances in coding, analysis and experimentation needed to support its wider use.
\end{abstract}

\noindent
Language agents plan, call tools, revise intermediate results and coordinate with other agents\cite{yao2023react,madaan2023self,shinn2023reflexion,zhu2025multiagentbench}. These activities leave trajectories: recorded sequences of messages, actions and observations. An agent may repeat an invalid tool call after corrective feedback; a revision may retain the error identified in the preceding critique. Understanding such behaviour requires learning how actions are organised across a trajectory and how that organisation relates to outcomes. This aim belongs to the broader study of machine behaviour\cite{rahwan2019machine,cherep2026behavioral,chen2025agentbehavioral,jackson2025aibehavioral} and its implications for human and social systems\cite{brinkmann2023machine,tsvetkova2024sociology,chan2023harms,kasirzadeh2026agentic}.

Task evaluation provides a starting point for this inquiry. Benchmarks assess performance\cite{liu2024agentbench,jimenez2024swebench,yao2025taubench,kwa2025measuring}, and AgentBoard tracks progress towards task completion\cite{ma2024agentboard}. Broader evaluation frameworks relate task demands to model capabilities\cite{zhou2026general}. Behavioural studies test specified actions such as self-correction, verification and backtracking\cite{huang2024cannot,kamoi2024when,gandhi2025cognitive}; MAST develops failure categories from interaction traces and investigates targeted changes\cite{cemri2025why}. An account of this process connects these measurements, interpreting each action in relation to preceding events and subsequent responses.

Such an account begins with identifying what to observe. Developing behavioural categories from recorded activity allows recurring patterns and unexpected responses to shape the analysis. Inductive coding offers one approach to developing these categories, or \emph{codes}\cite{glaser1967discovery,saldana2021coding}. Explicit definitions locate instances in a trajectory; their positions distinguish checking before an action from checking after an error. The resulting measurements retain the behavioural meanings needed to interpret relationships across the process.

\begin{figure}[!tb]
\centering
\includegraphics[width=\textwidth]{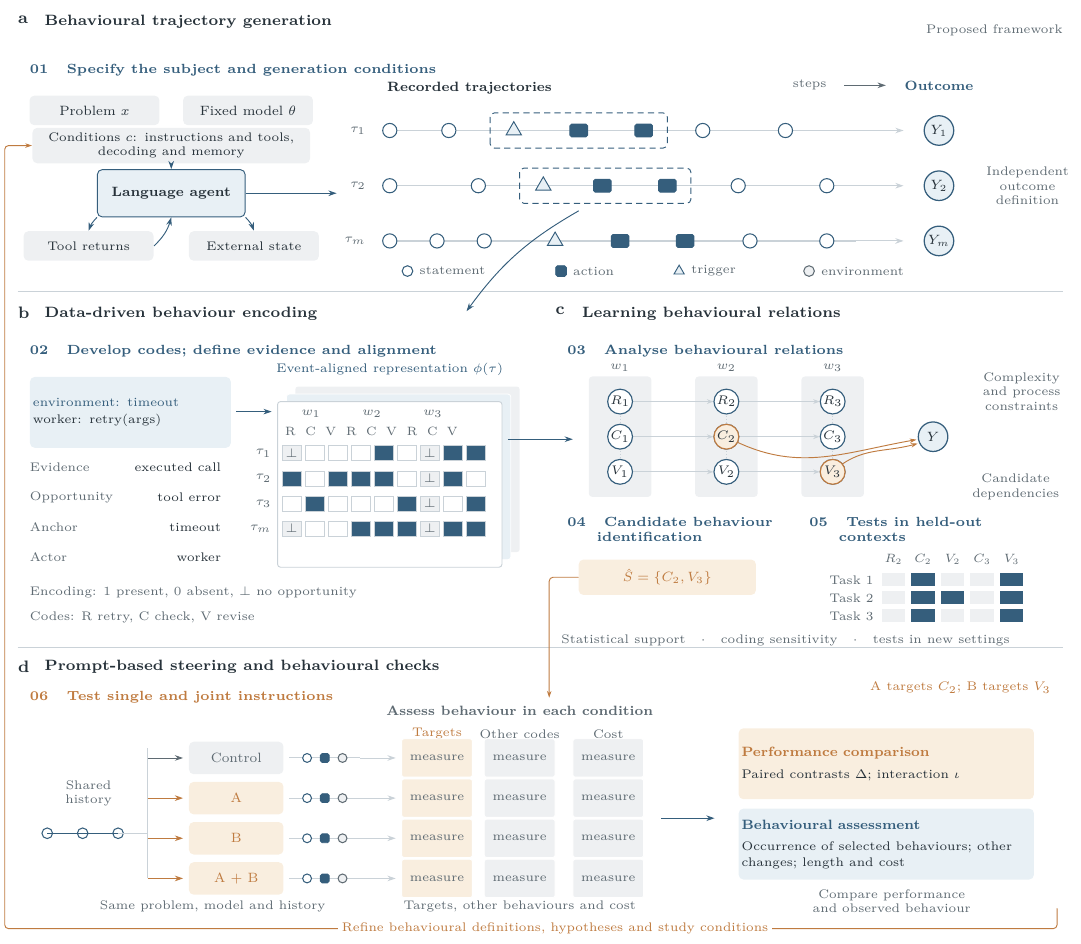}
\caption{\textbf{General framework for interpretable behaviour analysis.}
\textbf{a}, Generation conditions define the agent, its recorded trajectories and an independently specified outcome.
\textbf{b}, Categories induced from trajectory records are defined by explicit evidence, opportunities, anchors and actors; $\bot$ denotes no opportunity.
\textbf{c}, Analyses assess statistical support and generalisation, with safeguards against spurious associations. The graph illustrates one possible statistical representation. Highlighted variables are candidate targets; arrows denote dependencies. R, retry; C, check; V, revise; $w$, event-aligned window.
\textbf{d}, Continuations from the same recorded history compare baseline, single-target and joint instructions. Recoding measures targeted and other behaviours; outcomes and cost are recorded for each condition. Findings inform revisions to behavioural definitions, hypotheses and instructions. All matrices and trajectories are schematic.}
\label{fig:loop}
\end{figure}

We propose a general framework that connects behavioural discovery, analysis of trajectories and experimental investigation (Fig.~\ref{fig:loop}). It draws on the behavioural sciences' emphasis on observable activity and context\cite{tinbergen1963aims,anderson2014computational,baer1968dimensions,iwata1994functional,hanley2003functional,cooper2020applied}. Behavioural definitions connect the stages: the activity identified in a record is also the object of analysis and experimental measurement. Statistical assessment quantifies evidence for learned relationships; testing in held-out contexts examines how far they extend. These relationships guide experiments that vary instructions or interaction conditions and measure the resulting behaviours and outcomes. Analytical methods follow the behavioural question. Together, the stages turn recorded processes into interpretable accounts of behaviour that can be tested and refined.

Our work on ReBeCA\rebecaref{} provides a concrete illustration in iterative self-reflection. Behavioural descriptions proposed from trajectories and consolidated by human review became variables indexed by round. Causal discovery modelled their statistical dependencies\cite{spirtes2000causation,chickering2002optimal,andrews2020completeness}, identifying small behavioural sets that predicted outcomes better than models using all recorded variables. Their explicit meanings allowed the findings to inform single and combined instructions. Results varied across models and prompt combinations, illustrating the need to examine responses to instruction changes directly\cite{miehling2025evaluating}. ReBeCA thus connects behavioural discovery to interpretable hypotheses and experimental comparisons. We develop this logic for trajectories whose length, event order and agent roles vary.

\section{Learning behavioural variables from trajectories}
A behavioural account begins with categories that reflect the activity being studied. An \emph{episode} is one run of an agent on a problem; its trajectory records that run's messages, actions and observations. In ReBeCA, a language model proposed candidate behavioural descriptions, which human review consolidated\rebecaref{}. More systematic category development can draw on inductive coding\cite{glaser1967discovery,saldana2021coding} and language-model-assisted approaches\cite{ziems2024can,zhong2025hicode}. The aim is to let patterns in the records shape what enters the analysis.

The range of those records matters. Sampling diverse trajectories, including unusual responses, broadens the activity from which categories can emerge. Withholding outcomes during category development separates the identification of behaviour from its subsequent association with success. Comparing candidate categories against supporting and contrasting passages helps refine definitions, consolidate overlaps and uncover omissions. Separate development records provide a further test of coverage. Once the rules are fixed, independent trajectories allow assessment of per-category errors and human disagreement. These comparisons make the influence of sampling and interpretation open to inspection.

\begin{figure}[!tb]
\centering
\includegraphics[width=\textwidth]{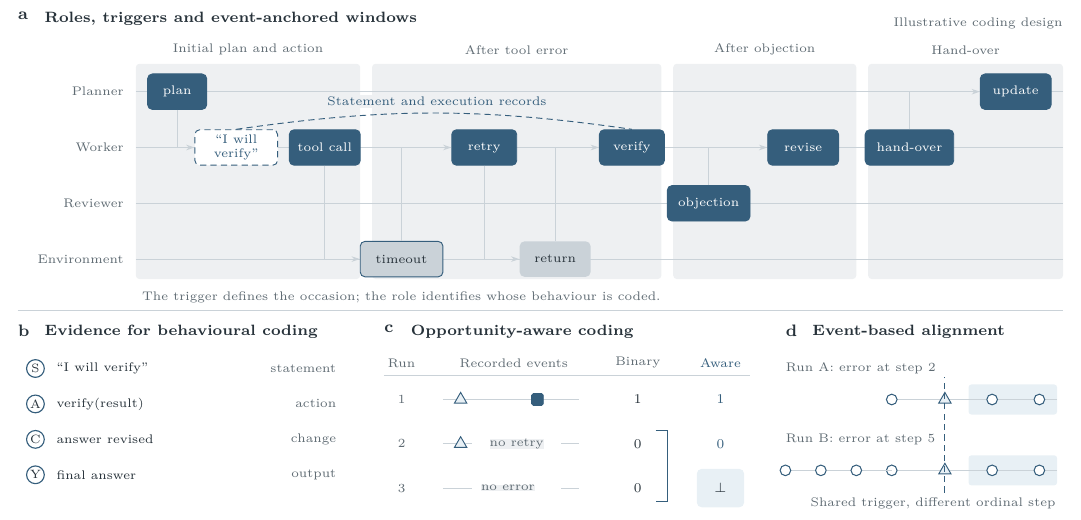}
\caption{\textbf{Measuring behaviour in context.}
\textbf{a}, An illustrative coding design includes three agent roles and an environment. Roles and event-anchored windows locate records of statements, executed actions and tool errors within the trajectory.
\textbf{b}, Statements, executed actions, subsequent changes and output properties provide distinct types of evidence.
\textbf{c}, A binary absence code merges an error without recovery with an episode without an error. Opportunity-aware coding uses $\bot$ for no opportunity; shared opportunities can themselves induce dependencies.
\textbf{d}, The same triggering event can occur at different ordinal steps. Event-aligned windows compare the corresponding occasions rather than a fixed step number. All examples are schematic.}
\label{fig:measurement}
\end{figure}

Category development leads to a measurement problem: the same behavioural description can refer to different activity depending on its evidence and context. Ethograms similarly define observable units of behaviour\cite{tinbergen1963aims,anderson2014computational,wiltschko2015mapping,datta2019computational,pereira2020quantifying}. For agents, preserving that meaning requires connecting each coded occurrence to the record, its occasion, its position and the actors involved (Fig.~\ref{fig:measurement}).

The evidence establishes what a code measures. ``I will verify the result'', a verification call, a subsequent revision and a correct final answer are distinct observations. Studies of verbal reasoning show why its correspondence with the model's computations warrants assessment\cite{turpin2023language,lanham2023measuring,chen2025reasoning,arcuschin2026chain,levy2026state}. A codebook can distinguish statements, executed actions and recorded state changes, and use counts, duration or latency where these answer the behavioural question (Fig.~\ref{fig:measurement}b).

An occurrence also has meaning relative to the opportunity for it. Recovery after a tool error requires an error; responding to an objection requires an objection. Distinguishing a response, absence despite an opportunity and no opportunity preserves this context. Shared opportunities can themselves induce dependencies. Representing the opportunity alongside the response allows both to be studied; restricting analysis to episodes where it arose instead concerns behaviour given that opportunity (Fig.~\ref{fig:measurement}c).

The same need for context extends to temporal position. Fixed reflection rounds provide common positions, whereas the fifth message of an open-ended episode can serve different purposes across runs. Windows anchored to a tool error, objection or hand-over offer a basis for comparison. Their interpretation depends on how absent and repeated events, overlapping windows and asynchronous actions are represented. Anchoring at the first occurrence gives one window per episode and confines the analysis to that occasion (Fig.~\ref{fig:measurement}d).

Within such a window, the trigger and the actor's role further specify the behaviour. A timeout and an invalid argument can call for different responses. An objection from a reviewer to a worker differs from one raised by a worker to a planner. Recording these distinctions, tool returns and external state lets a category retain the context in which an action occurred (Fig.~\ref{fig:measurement}a). Together, these choices turn descriptions into measurements of activity within a process.

The generation conditions give those measurements their empirical scope. With model parameters fixed and memory and external state reset, repeated episodes can be generated without learning from previous runs. Where the relevant state can be restored, alternative continuations can begin from the same history\cite{bogdan2025thought,lanham2023measuring}. Such comparisons depend on access to the agent's state and environment. Expectations of training or evaluation can also change behaviour\cite{greenblatt2024alignment,meinke2024frontier}. Recording these conditions makes it possible to relate learned patterns to the circumstances that produced them.

\section{Learning behavioural structure across a trajectory}
These measurements allow analysis to move from identifying behaviours to learning how they relate. A tool error may precede a retry, a change of tool or a revised plan; checking and revision may recur together. Indexing observations by round or event preserves where these relations arise. The behavioural question determines the analysis. Joint modelling, for example, examines which relationships remain informative given other recorded variables.

This conditional view can change the interpretation of an outcome association. Information carried by an early behaviour may already be represented by a later revision. Conditioning on the revision can then remove the earlier variable's additional predictive contribution. ReBeCA's \emph{pivotal behaviour identification} uses this logic to seek variables retaining outcome information given the other recorded variables\cite{pearl1988probabilistic,pellet2008using}. In its translation analysis, 11 of 15 code-derived variables were individually associated with the COMET outcome at an unadjusted $P<0.05$, while the selected set contained one such variable\rebecaref. The contrast illustrates how marginal association and conditional dependence answer different behavioural questions.

ReBeCA implements this analysis through causal discovery. Its Greedy Equivalence Search (GES) scores candidate graphs by data fit and model complexity\cite{chickering2002optimal}. BIC, for example, requires an improvement in fit sufficient to offset additional parameters. ReBeCA compares AIC, BIC and two Bayesian scores, while excluding backward-round edges and outgoing edges from the final outcome\rebecaref. Temporal restrictions encode background knowledge about admissible order\cite{andrews2020completeness}. Joint conditioning, complexity control and process constraints offer possible safeguards against redundant or spurious associations. Their interpretation rests on the recorded variables and modelling assumptions; independent assessment examines the resulting findings.

What can be learned also depends on how the process is represented. Process mining and causal discovery applied to agent--tool invocation records model workflow dependencies\cite{fournier2025observability}. CAMO combines computable factors from agent simulations with conditional structure learning and controlled simulator studies\cite{yu2026camo}. Invocation types, run-level factors and behaviours indexed by round or event describe different aspects of activity. Choosing among them follows the question and the behavioural meanings the analysis needs to preserve.

The outcome definition makes a further difference. Variables informative about final correctness may differ from those informative about recovery time or constraint violations; adding a revision code may change which earlier behaviours retain outcome information. Examining sensitivity to these choices shows how much a finding depends on its representation. A relationship with explicit behavioural meaning can then become a hypothesis for further study, including experiments that examine responses to instructions informed by that relationship.

\section{Interpretability from measurement to experiment}
Interpreting a learned relationship means relating it to identifiable activity. A conclusion specifies which behaviours are related, when, under what conditions and with what supporting evidence. If early checking and later revision jointly predict success, researchers should be able to locate both in the records and formulate a hypothesis about their relationship. Their meanings connect the statistical finding to activity that can be examined experimentally.

The same connection can make behavioural and internal analyses informative to one another. Internal analyses relate representations and computations to model outputs\cite{hewitt2019designing,belinkov2022probing,olah2020zoom,templeton2024scaling,lindsey2025biology,sharkey2025open}; high-level models of internal computation also offer an approach to interpretability\cite{geiger2025causal}. Relating these computations to defined behaviours would connect model mechanisms to the processes observed across a task. Neuroscientists have similarly advocated detailed behavioural analysis to inform mechanistic interpretation\cite{krakauer2017neuroscience,jonas2017microprocessor}.

Following a hypothesis through analysis and experiment requires distinguishing what each comparison establishes (Box~\ref{box:results}). In ReBeCA, marginal and conditional associations describe different aspects of the recorded relationships, while prediction assesses their information on new episodes. Instruction comparisons examine what happens when prompts change. Measuring behaviour under those conditions connects the experimental result to the hypothesis that motivated it. Supplementary Note~1 defines the corresponding quantities.

\begin{perspectivebox}{Box 1 $\mid$ Reading dependencies and steering results}
\label{box:results}
\small\sffamily
\begin{itemize}
\item \textbf{Marginal association, $A$.} Outcome differences between episodes with and without a coded behaviour. Other behaviours, problem difficulty and coding choices can contribute to the difference.
\item \textbf{Conditional information, $D$.} Additional outcome information given the other measured variables. The result depends on the recorded variables and outcome definition.
\item \textbf{Predictive assessment.} Prediction on new episodes, compared with models using no behavioural variables or all recorded variables.
\item \textbf{Instruction comparisons, $\Delta$ and $\iota$.} $\Delta$ measures performance change from baseline. $\iota$ compares the combined change with the sum of the two separate changes. All conditions use matched test problems. Recoding measures targeted and other behaviours.
\item \textbf{Representation constraints.} Shared opportunities and overlapping codes can create impossible value combinations. Joint encoding or opportunity-conditioned analysis changes what is represented and which question is answered.
\end{itemize}
\end{perspectivebox}

\Needspace{8\baselineskip}
\section{Testing behavioural hypotheses}
A learned relationship concerns the conditions in which its trajectories were collected. Significance tests, effect estimates and uncertainty intervals quantify evidence within that setting. For predictive analyses, comparison with simpler models shows what the selected behaviours add. ReBeCA uses cross-validated log-likelihood to compare selected sets with an outcome-only model and a model using all recorded variables (Supplementary Note~2)\cite{shmueli2010explain}. Separating selection from final assessment, through independent data or nested resampling, protects this comparison from the optimism introduced by choosing a candidate on its evaluation data.

Evidence within a setting raises the question of how far the relationship extends. Resampling examines whether selected variables recur across samples\cite{meinshausen2010stability}; varying coding choices and scoring rules tests sensitivity to the analysis. Assessment in contexts excluded from discovery, fitting and model selection examines generalisation across tasks, environments or generation conditions\cite{peters2016causal}. Random data splits address variation within the sampled setting. These assessments give a behavioural hypothesis an empirical scope that can guide its next test.

Experiments extend this inquiry by deliberately varying the conditions under which behaviour occurs. Explicit behavioural definitions make prompt steering one practical approach: a hypothesis informs an instruction, and the resulting trajectories show how the agent responds. Factorial designs compare single and combined instructions on new test problems, using the same problems and models in each condition\cite{fisher1935design}. Where state restoration is possible, continuations from the same history hold the preceding interaction fixed\cite{bogdan2025thought,lanham2023measuring}. The comparisons follow the logic of interventional research\cite{eberhardt2007interventions}; their estimates concern the tested prompt conditions and account for uncertainty, repeated observations and multiple comparisons.

Behavioural measurement is what connects these estimates back to the hypothesis. An instruction can change other behaviours, response length or cost alongside its intended target. Recoding selected and unselected behaviours identifies these accompanying changes. For an inconclusive performance contrast, the records help establish whether the target was elicited. For an improvement, they help identify changes for further tests. This evidence informs the next prompts, controls and hypotheses, allowing observation and experiment to refine one another.

\section{Lessons from the ReBeCA prototype}
ReBeCA\rebecaref{} brings these stages together in a fixed-round implementation of Self-Refine\cite{madaan2023self}, in which models critique and revise their own answers. Four Qwen3 models with 0.6--8 billion parameters\cite{yang2025qwen3} completed 417 mathematics episodes and 400 translation episodes, each with five rounds\cite{hendrycks2021math,andrews2025bouquet}. At each step, the most central response (medoid) in the largest semantic cluster among 20 sampled responses was retained. Outcomes were final-answer correctness for mathematics and change in translation quality under two model-based metrics\cite{rei2020comet,juraska2023metricx}.

\begin{figure}[!tb]
\centering
\includegraphics[width=\textwidth]{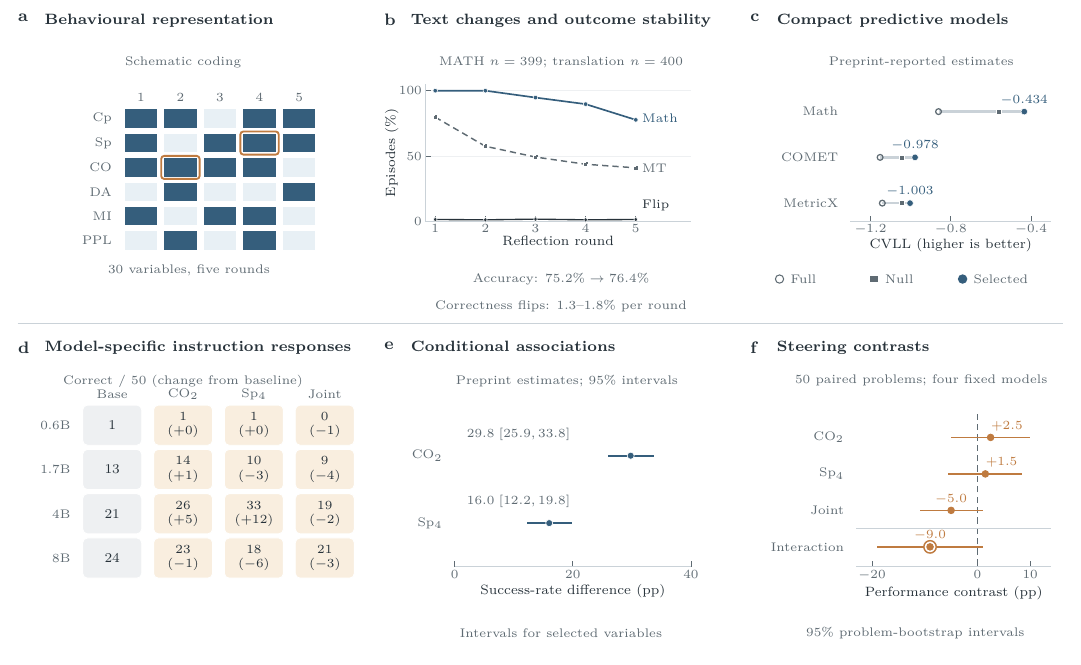}
\caption{\textbf{Measurement, prediction and instruction tests in ReBeCA.}
\textbf{a}, A schematic encoding matrix, generated for illustration, shows six variables over five rounds. Cp, completeness; Sp, specificity; CO, clarity and organization; DA, depth of analysis. $\mathrm{MI}$ and $\mathrm{PPL}$ indicate increases in estimated mutual information and perplexity; outlined cells mark selected round-specific variables.
\textbf{b}, Response-text changes in mathematics and translation (MT), and correctness flips in mathematics. Mathematics uses the same 399 fully graded episodes at every round; translation uses 400 trajectories. Initial and final mathematics accuracy are shown below.
\textbf{c}, Preprint-reported cross-validated log-likelihood (CVLL) for all variables (Full), no behavioural variables (Null) and selected variables; folds are defined by subject or source language. Connecting lines link estimates for each variable set.
\textbf{d}, Correct answers out of 50 AIME problems for each model and prompt condition; parentheses show count differences from that model's baseline. CO$_2$ and Sp$_4$ label instructions targeting clarity and organization at round two and specificity at round four; Joint combines them.
\textbf{e}, Preprint-reported conditional associations for $\co{2}$ and $\spc{4}$, averaged over 20 disjoint subsets, with 95\% confidence intervals unadjusted for selection.
\textbf{f}, Pooled steering contrasts relative to baseline and their interaction (open circle), in percentage points. Intervals use 20,000 paired bootstrap resamples of the 50 problems, retaining all four fixed models (seed 0). Panels b,d,f were recomputed from source outcomes; panels c,e reproduce preprint estimates.}
\label{fig:example}
\end{figure}

The trajectories expose a distinction between revision and improvement. In the 399 fully graded mathematics episodes, text changed in 77.7--100\% of episodes per round, while correctness flipped in only 1.3--1.8\%; initial and final accuracy were 75.2\% and 76.4\%. Learning what the revisions involved required behavioural measurements. Four behaviour codes proposed with outcomes withheld and consolidated by human review, together with two uncertainty indicators per round, yielded 30 variables per episode. The automated coder agreed with the majority label of three human coders on 89 of 100 instances. Fixed rounds supplied common positions and opportunities for critique (Fig.~\ref{fig:example}).

These measurements yielded small sets of outcome-relevant behaviours. The mathematics set contained clarity and organization at round two ($\co{2}$) and specificity at round four ($\spc{4}$); the translation set contained clarity and organization at round three ($\co{3}$). Both sets achieved better cross-validated predictive fit than the outcome-only and full-variable models (Fig.~\ref{fig:example}c). Conditional associations of $\co{2}$ and $\spc{4}$ with success averaged 29.8 and 16.0 percentage points across 20 disjoint subsets (Fig.~\ref{fig:example}e). Comparisons across four groups of subsets yielded no detectable differences; cross-environment transfer remains untested. Selection and comparison reused the same cross-validation, and associations were estimated from the selection data. The findings therefore identify candidates for further tests.

The selected dimensions then informed instructions targeting $\co{2}$, $\spc{4}$ or both on 50 AIME problems, using the same four models. Pooled correct-answer counts were 59 of 200 at baseline, 64 and 62 for the single instructions, and 49 for their combination. Baseline comparisons were non-significant after Holm correction across all 24 model-specific pairwise tests (Supplementary Note~3). The model-specific counts reveal differing responses (Fig.~\ref{fig:example}d): the specificity instruction solved 33 of 50 problems in the 4B model, against 21 at baseline, but 18 in the 8B model, against 24. In the 4B model, specificity alone outperformed the combined instruction (33 versus 19 correct). It solved 15 problems that the combined instruction missed; the reverse occurred on one problem (exact McNemar test, Holm-adjusted $P=0.012$). The pooled interaction estimate was $-9$ percentage points, with a problem-clustered 95\% bootstrap interval of $-19$ to $1$. These contrasts motivate further tests of the combination; the pooled interval remains compatible with zero interaction (Fig.~\ref{fig:example}f).

Interpreting these responses returns the analysis to measurement. The instructions restricted reflection to selected dimensions rather than simply adding the named behaviours. The coder saw reference outputs, agreement was pooled across codes, and each code combined the dimension evaluated in feedback with the dimension changed in revision. Medoid selection favoured common responses over rare ones. Recoding the steering trajectories, distinguishing feedback from revision and comparing alternative instructions for the same behaviours would clarify the findings. Transfer to other model families and less constrained tasks remains open. The prototype thus connects discovery with experiments while exposing the measurement and evaluation questions that a broader framework must address.

\Needspace{8\baselineskip}
\section{Extending the framework to open-ended behaviour}
The broader task is to preserve this connection as the organisation of activity changes. Planning, tool use and collaboration involve sequences of varying length and order (Table~\ref{tab:extension}). Repeated events, absent opportunities and actions involving several agents change what constitutes a comparable observation. Tool returns and external state help establish what happened\cite{yao2025taubench}; an initiator, recipient, executor and verifier may each contribute to a collaborative action\cite{zhu2025multiagentbench,cemri2025why}. Preserving these distinctions makes relationships across agents and events interpretable.

\begin{table}[!tb]
\centering\small\sffamily
\caption{\textbf{Behavioural questions beyond the prototype.} Proposed measurements distinguish actions, their opportunities and the questions they support. ReBeCA used a different codebook for self-reflection; the entries below specify proposed research designs.}
\label{tab:extension}
\begin{tabularx}{\textwidth}{@{}>{\raggedright\arraybackslash}p{0.155\textwidth}>{\raggedright\arraybackslash}p{0.26\textwidth}>{\raggedright\arraybackslash}p{0.2\textwidth}>{\raggedright\arraybackslash}X@{}}
\toprule
\textbf{Behaviour} & \textbf{Observations to distinguish} & \textbf{Opportunities} & \textbf{Relations to assess} \\
\midrule
Planning\cite{yao2023react} & Stated plan; precondition check; revised plan; executed step & New information; failed step & Replanning after new information; correspondence between stated plans and subsequent actions \\
Tool use\cite{yao2025taubench} & Tool choice; arguments; returned output; error response; state verification & Timeout; invalid argument; refused permission & Feedback preceding retries, tool changes and replanning; associations with recovery and cost \\
Self-reflection\rebecaref & Identified problem; proposed change; implemented change; renewed check & Critique; new evidence & Correspondence between critiques and revisions; variation across reflection rounds \\
Collaboration\cite{zhu2025multiagentbench,cemri2025why} & Work assignment; evidence request; objection; hand-over; shared-state update & Role-specific messages; transferred work & Interaction patterns associated with coordination, omission and duplicated work \\
\bottomrule
\end{tabularx}
\end{table}

For example, recovery after errors can be studied through the frequency of replanning after a first error and after repeated errors. Joint analysis can examine the relation given other recorded variables, and prompt comparisons can test instructions intended to change the response. Tracking replanning in both studies connects the observational finding to the experimental comparison. Recurrent and asynchronous events may call for sequence or event-based models that preserve the relevant behavioural meanings.

Where explicit symbolic actions and states are available, the process can also inform the analysis through learned action rules. A recommended extension of ReBeCA would combine Safe~Action~Model learning with causal-discovery-based structural dependency learning. Safe Action Model learning derives action preconditions and effects from observed transitions\cite{juba2021safe}. Within its planning model, next-state predictions are guaranteed correct for actions satisfying the learned preconditions. Mapping these rules to behavioural variables could constrain candidate dependencies before graph search. Hard exclusions require a justified impossibility rule; uncertain support can guide preferences among structures. Comparing rule-guided and unguided models on independent trajectories would assess the value of this extension.

\section{Building cumulative knowledge of agent behaviour}
Extending the framework across settings also creates an opportunity to connect findings. A relationship observed in one task becomes more informative when its behavioural meaning can be assessed in others. Such comparison depends on shared understanding of what was measured, how the trajectories arose and which evidence supported the conclusion (Box~\ref{box:reporting}).

\begin{perspectivebox}{Box 2 $\mid$ Reporting checklist for agent behaviour studies}
\label{box:reporting}
\small\sffamily
\begin{itemize}
\item \textbf{Agent and conditions.} Model, version and quantization; instructions, tools, decoding and stopping; state resets; sampling and selection.
\item \textbf{Trajectory and outcome.} Messages, actions, tool returns and state changes; rules for absent, repeated and overlapping events; outcome definition and metric, independent of the codebook.
\item \textbf{Codebook and coding.} Definitions, examples, evidence, opportunities, triggers and roles; induction and consolidation; coder access to references and outcomes; per-code agreement, prevalence and errors on an outcome-blinded sample.
\item \textbf{Analysis.} Question, methods and assumptions; effect estimates, significance and uncertainty; independent assessment and held-out contexts; appropriate comparison models; sensitivity to coding and setting.
\item \textbf{Experiment.} Prespecified single and combined instructions; matched or replay design; new test problems; performance differences, uncertainty and correction for multiple comparisons; changes in targeted and other behaviours, length and cost.
\item \textbf{Scope and provenance.} Model families, tasks and conditions; reused and new data; schematic and empirical displays; non-significant results and unsuccessful instruction combinations.
\end{itemize}
\end{perspectivebox}

Behavioural definitions provide the first connection across studies\cite{tinbergen1963aims,anderson2014computational,saldana2021coding}. Vocabularies induced from large conversation collections offer starting points\cite{tamkin2024clio,huang2025values}, while cross-model coding studies can examine where definitions transfer. Per-code prevalence, human disagreement and error types reveal where measurements need revision; automated coders also warrant checks for judge biases\cite{zheng2023judging}. Complete trajectory records let researchers revisit these judgements. Releasing them alongside scores\cite{burnell2023rethink,kapoor2025agents}, with tool returns, state changes, generation conditions and sampling rules, makes alternative analyses possible, subject to privacy and access constraints. Rare and exceptional events broaden the activity available for discovery.

This basis for comparison could connect behavioural questions now studied for different purposes. Codes that predict improvement under reinforcement learning\cite{gandhi2025cognitive} may also inform hypotheses about how behaviour changes during learning. Documented coordination failures suggest targets for experiments\cite{cemri2025why}. Safety studies generate evaluations\cite{perez2023discovering} and elicit deception, sandbagging and insider-threat behaviour\cite{meinke2024frontier,lynch2025agentic,marks2025auditing}; behavioural measurement can relate such findings to the recorded process. Verbal reasoning also offers a monitoring signal whose reliability requires assessment\cite{baker2025monitoring,korbak2025chain}. Internal interventions provide another way to examine hypotheses\cite{panickssery2024steering,venhoff2025understanding,geiger2025causal}, with appropriate controls informing their interpretation\cite{adebayo2018sanity,makelov2024subspace}.

Experiments can in turn improve the behavioural account from which they began. Single and combined prompts, graded instructions, repeated continuations and internal manipulations probe different aspects of a hypothesis\cite{panickssery2024steering,venhoff2025understanding}. Measuring targeted behaviours and accompanying changes connects their results. For graphical models, experimental observations can inform further structure learning when the manipulations and their targets are understood\cite{hauser2012characterization}. Comparing prompt formulations and evaluation contexts examines how far findings extend\cite{greenblatt2024alignment,meinke2024frontier}. Stating which observations support, revise or reject a hypothesis, and which populations the estimates describe, makes successful and unsuccessful tests informative to subsequent studies\cite{burnell2023rethink,weidinger2025evaluation}.

Extending this cycle brings new research questions. A category developed in one model or language may need revision in another; comparing its instances can reveal which behavioural distinctions persist. Recurring and asynchronous actions raise a related problem of representing interactions while preserving their meanings. Experimental findings can inform these choices, and adaptive designs could select tests that distinguish competing hypotheses. Automated discovery and experimentation\cite{king2009automation,wang2023scientific,gottweis2026accelerating,liu2024discovery,lee2026automating,yu2026camo} could help sustain this exchange between behavioural observation and model refinement.

\Needspace{10\baselineskip}
\section{Outlook}
The framework connects discovery, analysis and experiment through explicit behavioural meanings, allowing findings to be tested and refined across contexts. ReBeCA illustrates one implementation in self-reflection. Its broader promise is an empirical account of how agents respond to information, coordinate their actions and reach outcomes.

\par
\begingroup
\small
\setstretch{1}
\titlespacing*{\section}{0pt}{10pt}{4pt}
\section*{Acknowledgements}
This work received no specific funding. 

AI tools such as OpenAI's ChatGPT and Anthropic's Claude assisted with language refinement of the paper, programming, and part of the verification process.


\section*{Author contributions}
T.Y. conceived the Perspective, developed the conceptual framework, performed the analyses, prepared the figures and wrote the manuscript. L.Q. provided theoretical and methodological guidance and research support. All authors revised the manuscript and approved the final version.

\section*{Competing interests}
The authors declare no competing interests.

\section*{Data availability}
The self-reflection study is described in the public ReBeCA preprint\rebecaref. Source data were used to recompute trajectory summaries and instruction-test statistics. Derived numerical summaries accompany this Perspective. The underlying trajectories and instruction-test data will be deposited in a public repository on publication.

\section*{Code availability}
The scripts used to verify the supporting summaries are available to editors and reviewers upon request for peer review.

\endgroup
\clearpage
\begingroup
\small
\setstretch{1}
\bibliographystyle{unsrtnat}
\bibliography{refs}
\endgroup

\clearpage
\section*{Supplementary Information}
\renewcommand{\theequation}{S\arabic{equation}}
\setcounter{equation}{0}
\subsection*{Supplementary Note 1 $\mid$ Behavioural variables and comparison quantities}
The definitions below illustrate the statistical comparisons used in the Perspective. An agent with fixed parameters $\theta$ generates an episode $\tau$ on a problem $x$ under conditions $c$, including its instructions, tools, decoding and stopping rule. A codebook gives each behavioural category a definition and a rule linking it to the trajectory record. In a fixed-round study, indexing the codes and numerical indicators by round yields a finite set of variables $V$. Event-anchored studies would require a corresponding rule for recurring or absent events. A code may indicate presence, frequency or duration. Where opportunity matters, it can distinguish absence despite an opportunity from no opportunity.

Let $Y(x,\tau)$ be an outcome defined independently of the codebook, and $P_c$ the joint distribution of $(V,Y)$ under the specified conditions. An instruction change $\eta$ produces conditions $c^{\eta}$, and $Q$ is a prespecified distribution of test problems.

\textbf{Observational and steering quantities.} The association $A(v)$ compares outcomes when a binary behaviour $v$ is present or absent. Conditional dependence $D(v)$ measures the information that $v$ adds given the other variables. The performance contrast $\Delta(\eta)$ compares expected outcomes under the original and modified instructions. The interaction $\iota$ measures departure from additivity when two instruction changes are combined:
\begin{equation}
\begin{aligned}
A(v)&=P_c(Y{=}1\mid v{=}1)-P_c(Y{=}1\mid v{=}0),\\
D(v)&=I\big(Y;\,v\mid V\setminus\{v\}\big),\\
\Delta(\eta)&=\mathbb{E}_{x\sim Q}\Big[\mathbb{E}_{\tau\sim P_\theta(\cdot\mid x,c^{\eta})}Y(x,\tau)-\mathbb{E}_{\tau\sim P_\theta(\cdot\mid x,c)}Y(x,\tau)\Big],\\
\iota(\eta,\eta')&=\Delta(\eta\eta')-\Delta(\eta)-\Delta(\eta').
\end{aligned}
\label{eq:defs}
\end{equation}
Here $I$ is conditional mutual information and $\eta\eta'$ denotes the combined instruction. The definition of $A$ assumes binary $v$ and $Y$; $D$ also applies to multivalued variables. Unlike $A$ and $D$, which describe observed episodes, $\Delta$ and $\iota$ compare experimental conditions. They summarize the implemented prompt conditions, which may differ in several aspects of generated behaviour.

These definitions distinguish observational and steering comparisons. Under strict positivity, variables with $D>0$ form the unique Markov boundary, the minimal set carrying all recorded predictive information about $Y$\cite{pearl1988probabilistic,pellet2008using}. Finite-sample search estimates this population target. The contrasts $\Delta$ and $\iota$ compare instructions on a specified test population. Estimating those contrasts requires observations under the corresponding prompt conditions.

\subsection*{Supplementary Note 2 $\mid$ Predictive assessment}
For a candidate variable set $M\subseteq V$, cross-validated log-likelihood measures predictive fit on data excluded from model fitting:
\begin{equation}
\mathrm{CVLL}(M)=\frac{1}{F}\sum_{f=1}^{F}\frac{1}{|\mathcal{D}_f^{\mathrm{test}}|}\sum_{i\in \mathcal{D}_f^{\mathrm{test}}}\log \hat p_f\big(y_i\mid M_i;\,\mathcal{D}_f^{\mathrm{train}}\big),
\label{eq:cvll}
\end{equation}
where $F$ is the number of folds, $\hat p_f$ is fitted on the data outside test fold $\mathcal{D}_f^{\mathrm{test}}$, and $M_i$ gives the values of the selected variables in episode $i$. The empty set predicts outcomes without behavioural variables; $V$ uses all recorded variables. Higher values indicate better predictive fit under the specified estimator and fold construction. Equal averaging across folds weights contexts equally when folds represent contexts; it differs from pooling episode-level log-likelihoods when fold sizes vary. The selected set's score requires independent or nested assessment if the same criterion was used to choose it.

\subsection*{Supplementary Note 3 $\mid$ Verification of supporting summaries}
The supplied source archive contains 418 mathematics records: 417 have the initial answer and five revisions, and one has only five outputs. Of the 417 complete-output episodes, 399 have non-missing grades at every position; Fig.~\ref{fig:example}b uses this subset throughout. The two translation-metric exports contain the same 400 trajectories, assessed under two metrics. The plot omits a legacy boxed-expression comparison, which conflates answer changes with extraction failures and syntactic differences. The verification output retains this diagnostic for transparency.

The instruction-test CSV contains 50 distinct problems and binary outcomes for four models under four conditions. Bootstrap resampling selects problems with replacement, retaining all model and condition outcomes within each problem; the same four models are retained in every resample. The 95\% percentile intervals use 20,000 resamples, NumPy's default random generator with seed 0 and linear percentile interpolation. Exact two-sided McNemar tests use the discordant pairs, with a single Holm correction across the 24 tests formed by six condition pairs in each of four models. Cochran's $Q$ uses its asymptotic chi-squared reference with three degrees of freedom, with Holm correction over four models. Two implementations of the paired problem-level bootstrap produced the same contrasts and intervals. Code-level annotations and fitted fold models are absent from this archive; the CVLL and coefficient summaries in Fig.~\ref{fig:example}c,e remain attributed to the ReBeCA preprint as reported estimates. Source-file hashes and numerical verification summaries accompany the manuscript. Verification scripts are available to editors and reviewers upon request.

\end{document}